\documentclass[letterpaper, 10 pt, conference]{ieeeconf}  

\IEEEoverridecommandlockouts                              

\usepackage{graphics} 
\usepackage{graphicx}
\usepackage{amsmath} 
\usepackage{makecell}

\usepackage{xcolor}
\usepackage{soul}
\usepackage{booktabs} 
\usepackage{float}
\usepackage[absolute,overlay]{textpos}

\title{\LARGE \bf
SHAFT: A Slack-Compensating, Helical-Buckling-Attenuating Flexible-Shaft Transmission for Lightweight Multi-DoF Manipulation
}

\author{Tomoya Takahashi$^{\dagger, 1}$, Moses Gladson Selvamuthu$^{\dagger, 1^*, 2}$ , Richiro Tadakuma$^{2}$, and Kazutoshi Tanaka$^{1}$
\thanks{${\dagger}$Equal Contribution}
\thanks{$^{*}$ Work done during internships at OMRON SINIC X.}
\thanks{$^{1}$OMRON SINIC X Corporation, Hongo 5-24-5, Bunkyo-ku, Tokyo, Japan.{\tt\small tomoya.takahashi@sinicx.com}}%
\thanks{$^{2}$Yamagata University, 4 Chome-3-16 Jonan, Yonezawa, Japan.
    {\tt\small mosesgladsonrocks@gmail.com}}%
\thanks{This work was supported by JSPS KAKENHI Grant Number JP25H01159, Japan.}
}
\begin{document}

\maketitle
\thispagestyle{empty}
\pagestyle{empty}

\begin{textblock*}{18cm}(1.5cm,0.5cm) 
    \tiny 2026 IEEE/RSJ International Conference on Intelligent Robots and Systems (IROS2026). Preprint. Accepted June 2026. \textcopyright 2026 IEEE.  Personal use of this material is permitted.  Permission from IEEE must be obtained for all other uses, in any current or future media, including reprinting/republishing this material for advertising or promotional purposes, creating new collective works, for resale or redistribution to servers or lists, or reuse of any copyrighted component of this work in other works.
\end{textblock*}

\begin{abstract}
Lightweight and slim manipulators enable safe operation in human living environments. Proximal actuation using remote transmission mechanisms, such as wire-driven or Bowden cables, effectively reduces inertia and arm size by relocating motors near the base and transmitting torque to distal joints. Existing approaches either increase mass through additional components, such as pulleys for direction changes, or suffer from reduced transmission efficiency due to friction losses. Flexible shaft transmission avoids both mass increase and excessive friction losses, but faces increasing angular transmission error due to helical buckling caused by slack generated at joint bending. 
To address this problem, we propose SHAFT: a Slack-compensating, Helical-buckling-Attenuating Flexible-shaft Transmission mechanism. This mechanism compensates for slack through a proximal tensioner, improving the angular transmission error and efficiency of flexible shaft transmission without increasing the moving mass of the arm section. In a transmission path containing four 90$^{\circ}$ bends, the proposed mechanism demonstrated approximately 30\% higher efficiency and approximately 65\% lower angular transmission error compared to a flexible shaft transmission without a tensioner. Using this mechanism, we fabricated a 6-Degree-of-Freedom (DoF) arm with a 1-DoF gripper manipulator consisting of a rotary module housing motors with a tensioner, and a 280~g weight of the arm module.
The proposed manipulator represents a novel remote actuation system for achieving lightweight construction with high efficiency, contributing to the acceleration of safe robot deployment in human environments.
\end{abstract}

\section{INTRODUCTION}

Manipulators operating in human-centered environments require lightweight, slim construction to ensure safe task execution in unexpected collisions with humans or objects. Reducing the moving mass of a manipulator decreases its inertia and overall weight, thereby minimizing potential impact forces and enhancing safety during collaborative tasks. A slimmer manipulator reduces the risk of unintended contact in cluttered environments. 

A well-established approach to reducing the mass of the arm section, the movable links of the manipulator whose weight directly loads the actuators, is proximal actuation, 
where motors are positioned near the base and torque is transmitted to 
the joints through remote transmission mechanisms. Various remote transmission methods have been proposed, including wire-driven systems with cable-pulley arrangements~\cite{endo2019super,tanaka2023twist}, pneumatic actuation in continuum arms, and mechanical linkages~\cite{shahhosseini2016new}.

\begin{figure}[t]
  \centering
  \includegraphics[width=1.0\linewidth]{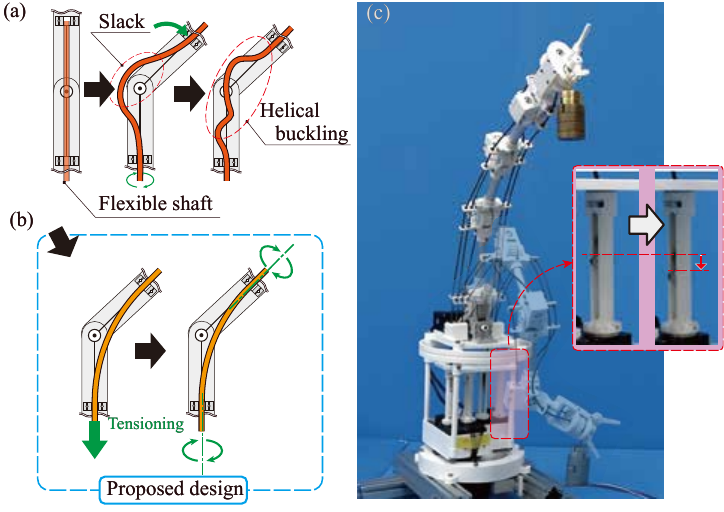}
  \caption{(a)~Helical buckling of flexible shafts caused by slack, (b)~Proposed method: Slack compensation suppresses angular transmission error, (c)~Fabricated flexible shaft-based manipulator.}
  \label{fig_teaser}
\end{figure}

We aim to realize a multi-degree-of-freedom (DoF) articulated manipulator capable of large joint bending angles to improve maneuverability in confined spaces, as in conventional joint-mounted motor designs. However, achieving this with remote transmission becomes challenging when the transmission path must traverse multiple joints with sharp bends.
Conventional linkages and wire-driven systems therefore require direction-changing mechanisms such as bevel gears or pulleys at each joint, increasing weight and mechanical complexity. Bowden cables, comprising an inner wire routed through a flexible sheath and transmitting tension, offer an advantage in that their flexible sheaths can follow arbitrary curved paths, but sliding friction between the cable and sheath increases with bending angle, resulting in significant transmission losses and reduced efficiency~\cite{jeong2015feedforward}. Consequently, the transmission efficiency, the ratio of output torque to input torque, degrades significantly depending on the manipulator's configuration.

We focus on flexible shaft transmission, a helically wound wire that transmits rotational torque along a curved path, as a promising alternative that addresses the limitations of both approaches. Like Bowden cables, flexible shafts can follow arbitrary path geometries without requiring direction-changing mechanisms at the joints. Unlike Bowden cables, however, the friction in flexible shaft transmission does not increase exponentially with bending angle, as the shaft is supported by discrete bearings that rely primarily on rolling friction rather than sliding friction. These properties make flexible shafts an attractive candidate for lightweight, high-efficiency remote actuation in multi-joint manipulators.  

However, flexible shaft transmission exhibits an increase in angular transmission error---the angular error between motor input rotation and joint output rotation---as the bending angle increases. 
When a joint bends, slack develops in the shaft path between bearing supports. Under torque loading, this slack allows the shaft to undergo helical buckling, a phenomenon where the shaft deforms helically rather than maintaining a planar curved trajectory (Fig.~\ref{fig_teaser}~(a)), resulting in elastic deformation that manifests as angular transmission error. One existing design solution is to cover the flexible shaft with a conduit~\cite{usman2024design}, which suppresses helical buckling by constraining the shaft path. However, this approach introduces sliding friction across the entire shaft surface, increasing both friction losses and weight of arm section.

In contrast to existing solutions, we propose \textbf{SHAFT} (\textbf{S}lack compensating, \textbf{H}elical buckling \textbf{A}ttenuating \textbf{F}lexible shaft \textbf{T}ransmission) mechanism to deliberately constrain the shaft path geometry, despite the inherent flexibility that allows arbitrary routing, to passively minimize slack~(Fig.~\ref{fig_teaser}~(b)), suppressing helical buckling without the friction penalties of conduits. Specifically, we employ a spring-based tensioner mounted at the manipulator base that applies axial tension to the flexible shafts (Fig.~\ref{fig_teaser}~(c)). As joints articulate, the tensioner passively minimizes slack by adjusting the effective shaft path length, maintaining the shortest possible path across all joint configurations and thereby suppressing helical buckling. Since the tensioner is located at the base and not at the joints, moving mass remains unchanged, and the lightweight remote transmission mechanism consists only of discrete bearings for shaft support.

Our design offers an enhanced approach to balancing lightweight construction, high transmission efficiency, and low angular transmission error in remote actuation systems.
The contributions of this paper are as follows:

\begin{enumerate}
    \item A passive base-mounted tension mechanism for flexible shaft transmission that minimizes slack across all joint configurations, suppressing helical buckling without the friction penalties of conduits or the weight penalties of joint-level mechanisms
    \item Experimental characterization of transmission efficiency and angular transmission error, demonstrating quantitative improvements over conventional flexible shaft designs
    \item A complete 6-DoF manipulator design with integrated 1-DoF gripper, validating the practical feasibility of the proposed approach
\end{enumerate}



\section{Related Work}

\subsection{Lightweight Manipulators with Remote Actuation}

We categorize remote transmission methods into two types: \textit{discrete-point redirection}, where mechanical components enable localized direction changes at distributed points, and \textit{continuous-path routing}, where flexible conduits enable smooth path design along continuous curves. The former excels in accurate force transmission with low friction losses, while the latter offers lightweight construction by eliminating direction-changing mechanisms at joints.


\subsubsection{Discrete-Point Redirection}

Discrete-point redirection has been widely adopted in robotic arms, including rigid-linkage designs using bevel gears~\cite{shahhosseini2016new} and wire-driven systems routing wires via pulleys at each joint~\cite{tanaka2023twist, endo2019super}. Various cable-driven manipulators have been proposed with optimized architectures featuring tension-amplification and dedicated tensioning mechanisms~\cite{pang2022stiffness, huang2020novel, yan2025lightweight}. Luo et al.~\cite{LuoD3arm} further demonstrated decoupled wire-driven actuation through double-joint mechanisms, enabling independent control of joint motions.


These discrete-point redirection approaches enable slim-profile designs in which transmission paths closely follow the link geometry and have relatively low friction losses at bending points due to rotating components such as bearings. However, they require mechanical components such as bevel gears or pulleys for direction changes at each joint, resulting in complex and heavy mechanisms.

\subsubsection{Continuous-Path Routing}

Bowden cable systems consist of a flexible inner cable enclosed within a flexible outer sheath, enabling tension transmission with only two fixed points on the outer tube~\cite{wang2023modular, yin2020tendonsheath, marais2017cram}. The key advantage is that the flexible sheath can follow arbitrary routing geometries, eliminating the need for direction-changing mechanisms at the joints. 

Continuum manipulators~\cite{mishra2017simba, dong2025bioinspired} achieve smooth bending by routing wires along their continuously deformable structure, but face challenges in achieving large curvature bending at discrete joints. Pneumatic actuation, such as the 7-DoF arm driven by artificial muscles~\cite{tondu2005seven}, offers extremely lightweight solutions by transmitting compressed air through tubes to joints. 

Continuous-path routing methods offer lightweight construction and high design freedom. However, they suffer from efficiency losses due to tube friction or pressure, and their path geometry depends on conduit elasticity, making it difficult to bend large curvatures at manipulator joints.



\subsection{Flexible Shaft Manipulator Designs}

Usman et al.~\cite{usman2024design} proposed three prototypes of manipulators using flexible shafts. The flexible shaft without conduit employed bearing-supported shafts but experienced significant helical buckling effects as bending angles increased. The flexible shaft with conduit and planar links incorporated flexible conduits covering the shafts, which reduced helical buckling but increased friction and required multiple support points, resulting in increased weight. The flexible shaft with conduit and cylindrical links combined flexible shaft segments with metal pipe, minimizing and optimizing the deformable sections to address these issues. While this work successfully demonstrated high-torque transmission through flexible shafts, only the elbow joint was actuated by flexible shaft transmission in their implementation~\cite{usman2024flexible}. The weight of each flexible shaft unit becomes particularly problematic when driving multiple joints with flexible shafts.

The inherent angular compliance of flexible shafts has been shown to function similarly to series elastic actuators (SEA), providing benefits for force control applications~\cite{choi2025flexi}. Flexible shaft mechanisms combining both tension and torque transmission have also been proposed in continuum-arm configurations, finding applications in medical miniature manipulators~\cite{sekiguchi2010development, liu2013flexible}. 
Similarly, flexible shafts have been shown to enable lightweight remote actuation in crawler-type rescue robots~\cite{hayashi2010torque}. 
Flexible shafts have been integrated into 3D-printed soft continuum actuators to realize push-pull-twist actuation and enhanced dexterity~\cite{tan2018design}. 

Among existing flexible-shaft manipulators for articulated multi-DoF arms, no mechanisms have been proposed to reduce angular transmission error without using outer tubes while maintaining low moving mass. As reported by Usman et al.~\cite{usman2024design}, shaft slack at joints can cause helical buckling under high torque and large bending angles. The common solution is to add outer tubes or cylindrical links to stiffen the shaft. In contrast, the base-mounted tensioner proposed in this study improves angular transmission error while minimizing additional components and moving mass, representing a novel approach in the literature.

\section{Method}

\subsection{Basic Principle of Flexible shaft Transmission}

\subsubsection{Slack Geometry Model}

This subsection presents a geometric model for the amount of slack generated with respect to the joint bending angle, and demonstrates the calculation method for the required tensioning distance at the base linear tensioner. The model aims to explain the onset mechanism of helical buckling as an initial consideration. Although the flexible shaft exhibits nonlinear torsional compliance even in its straight state, this effect is assumed to be sufficiently small relative to helical buckling.
\begin{figure}[t]
  \centering
  \includegraphics[width=0.75\linewidth]{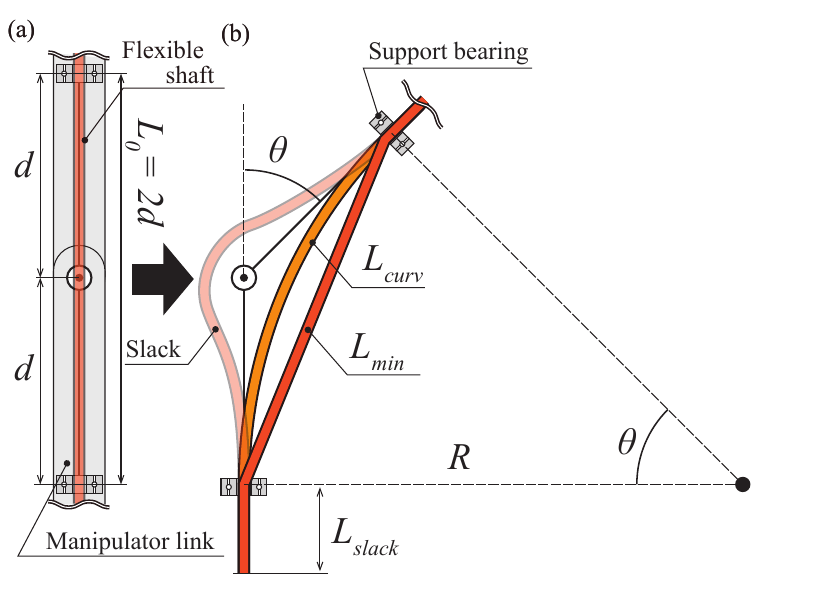}
  \caption{(a)~straight position of link, (b)~Constant curvature shape of flexible shaft.}
  \label{fig_slack_geometry}
\end{figure}

We first derive the slack generation as a function of joint bending angle. Consider support bearings placed at a distance $d$ from the joint rotation axis on both sides, as shown in Fig.~\ref{fig_slack_geometry}~(a). In the straight configuration, the shaft length between bearings is $2d$. When the joint is bent by angle $\theta$, assuming the shaft follows a constant curvature path between the bearings~(Fig.~\ref{fig_slack_geometry}~(b)), the arc length becomes

\begin{equation}
L_{\text{curve}} = R\theta = \frac{d}{\tan(\frac{\theta}{2})}\theta
\end{equation}

where $R = d/\tan(\theta/2)$ is the radius of curvature. Note that $L_{\text{curve}}$ equals $2d$ when $\theta = 0$ and decreases as $\theta$ increases. Therefore, the slack length is given by

\begin{equation}
L_{\text{slack}} = 2d - \frac{d}{\tan(\frac{\theta}{2})}\theta
\end{equation}

For example, assuming a maximum joint bending angle of~90$^{\circ}$, the slack becomes 
\begin{equation}
L_{\text{slack}} = 2d - d\frac{\pi}{2} = \left(2 - \frac{\pi}{2}\right)d
\end{equation}

Without axial sliding compensation, this slack causes deformation involving multiple curvatures. When the flexible shaft spans $N$ joints, each bent at the maximum angle, the total required tensioning distance is

\begin{equation}
L_{\text{total}} = N \cdot L_{\text{slack}} = N\left(2 - \frac{\pi}{2}\right)d
\end{equation}

This value represents the minimum extension length necessary to reduce slack and ensure stable torque transmission through the flexible shaft system (Fig.~\ref{fig_slack_geometry}~(b)). Based on the geometric configuration of the joint, the minimum shaft length corresponding to the diagonal configuration introduced by the tensioner can be expressed as

\begin{equation}
L_{\min} = 2d - \sqrt{2}d
\end{equation}

The maximum extension for this configuration is given by 

\begin{equation}
L_{\max} = N(2-\sqrt{2})d
\end{equation}

\subsection{System Design}
\subsubsection{Flexible Shaft Transmission Unit}

This section describes the specific design of the flexible shaft unit.

\begin{figure}[t]
  \centering
  \includegraphics[width=0.9\linewidth]
  {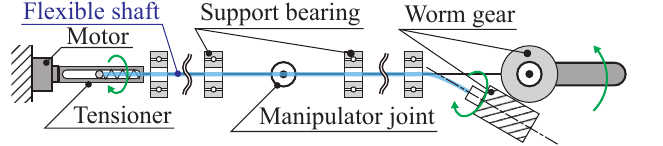}
  \caption{Schematic of single remote actuation unit of SHAFT.}
  \label{fig_single_module}
\end{figure}

Fig.~\ref{fig_single_module} shows the design of the remote actuation unit, which drives the manipulator links through a servo motor, tensioner, shaft, bearings, and reduction mechanism. For all joint actuation, servomotors (Dynamixel XM540-W270-R, ROBOTIS) are employed. The joint torque is transmitted through a commercially available 3~mm flexible rotary shaft made of high-carbon steel. The shaft consists of a multi-layer helically wound steel wire about a high-carbon spring steel core, providing high torsional flexibility while maintaining adequate torque transmission capability.  

\begin{figure}[t]
  \centering
  \includegraphics[width=0.7\linewidth]
  {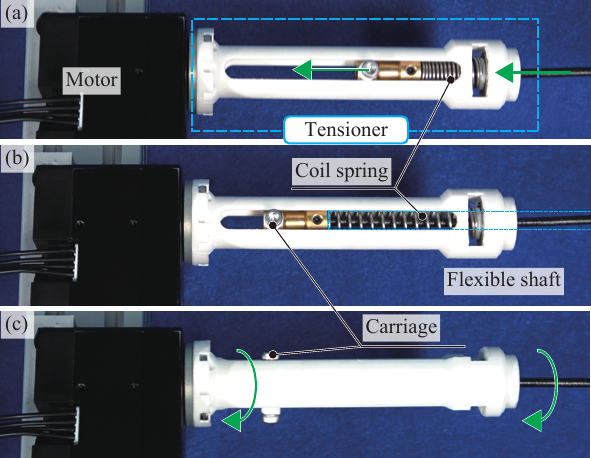}
  \caption{Tensioner mechanism: (a)~Released position, (b)~Contracted position, (c)~Rotation motion.}
  \label{fig_tensioner}
\end{figure}

The tensioning mechanism consists of a slotted tube integrated with the motor unit~(Fig.~\ref{fig_tensioner}). All components of the tensioner assembly were fabricated using 3D-printed PLA. Inside this tube, a movable carriage that slides passively is installed. The carriage is a brass coupler, which is rigidly fixed to the terminal end of the 3~mm flexible shaft. 
A set of two compression springs, with an effective combined stiffness of 0.2~N/mm and a working length of 80~mm, is placed inside the slotted tube and preloaded during assembly. 
The working length and preload were selected such that sufficient axial tension is maintained to prevent slack generation across the full range of joint bending angles from 0$^{\circ}$ to 90$^{\circ}$.
The spring continuously applies an axial force to the carriage, thereby maintaining constant tension in the flexible shaft. When the manipulator joints bend and the effective shaft length changes, the carriage moves passively along the slotted tube. The constant spring force ensures smooth sliding motion and prevents the formation of slack, enabling stable and reliable power transmission.

The flexible shaft is routed along the manipulator through intermediate support bearings with an inner diameter of 4~mm fixed on the links. These bearings provide radial support and allow the shaft to rotate freely together with the inner race, ensuring smooth torque transmission. At the same time, the small shaft-bearing clearance allows passive axial sliding while maintaining alignment and stable torque transmission to distal joints.


\subsubsection{Manipulator Design}

\begin{figure}[t]
  \centering
  \includegraphics[width=0.7\linewidth]{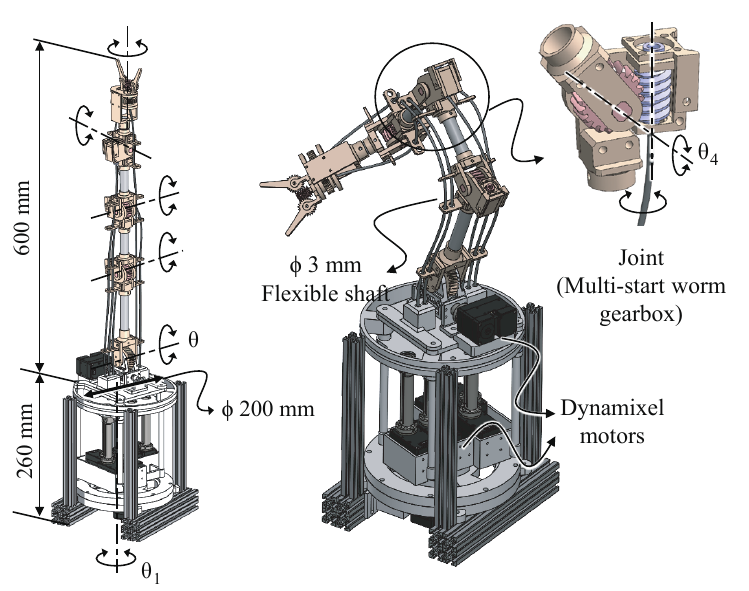}
  \caption{Basic design of manipulator driven by SHAFT.}
  \label{fig_basic_design}
\end{figure}

\begin{figure}[t]
  \centering
  \includegraphics[width=0.7\linewidth]{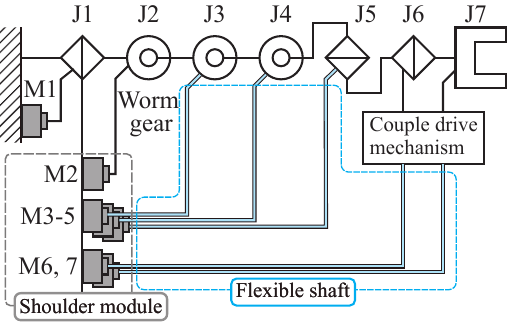}
  \caption{Schematic of joints, motors, and flexible shaft configuration of manipulator.}
  \label{fig_joint_configuration}
\end{figure}

We designed the proposed transmission mechanism as a 6-DoF arm with a 1-DoF gripper~(Fig.~\ref{fig_basic_design}). The manipulator consists of the shoulder module and the arm section. The shoulder module is a structure rotated by J1 and contains the motors for J2 and the motors/tensioners for J3–J7. Since this module rotates about the vertical yaw axis, its mass does not impose gravitational loads on the J1 motor. The arm section comprises joints J3-J7 along with their flexible shafts and joint reduction mechanisms. Minimizing the mass of this arm module directly reduces the load on the joint drive motors. 
As shown in Fig.~\ref{fig_joint_configuration}, the proposed transmission mechanism using flexible shafts actuates joints J3--J7, while servo motors drive shoulder yaw (J1) directly and shoulder pitch (J2) through reduction gears. Joints J6 and the gripper (J7) employ a coupled drive mechanism. All motors are controlled via a serial converter connected to a PC. The desired end-effector position is specified by the user, and the corresponding joint angles are computed using inverse kinematics to drive the manipulator.

All structural components were fabricated using 3D-printed PLA to achieve a lightweight design. The links between joints were constructed from aluminium tubes to improve stiffness while maintaining low mass. To guide the flexible shaft along each link, two support bearings were installed at a distance of $d=25$~mm from the joint axis. The manipulator has an overall length of approximately 600~mm, with a shoulder module height of 260~mm and a diameter of 200~mm. The shoulder module weighs 1650~g, while the arm section weighs 280~g. The fabricated 6-DoF manipulator with a 1-DoF gripper is shown in Fig.~\ref{fig_teaser}~(c). The following describes the detailed design of each section of the manipulator.

\begin{figure}[t]
  \centering
  \includegraphics[width=0.8\linewidth]
  {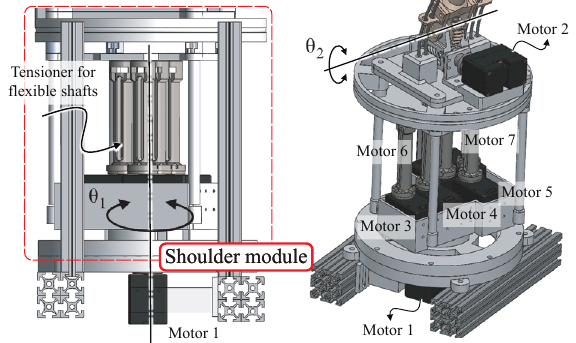}
  \caption{Design of shoulder module and J1 motor.}
  \label{fig_base_module}
\end{figure}

\textbf{Shoulder Module and J2}: 
The shoulder module consists of a cylindrical structure that rotates in the yaw direction, directly driven by the J1 motor as shown in Fig.~\ref{fig_base_module}. Inside the shoulder module, motors and tensioners for joints J3 through J7 are housed and connected to the flexible shafts that drive each joint. The top of the shoulder module contains the J2 joint, and a servo motor is connected to the joint through a 10:1 multi-start worm gearbox.

\textbf{J3 to J5}: 
Joints J3--J5 are driven by flexible shafts of different lengths, each supported by two bearings per link. These joints employ worm gear reductions of 20:1, except for J5, which uses a 10:1 reduction, as shown in Fig.~\ref{fig_teaser}. The reduction increases payload capacity while reducing shaft twist and improving overall stiffness during operation.

\begin{figure}[t]
  \centering
  \includegraphics[width=0.75\linewidth]
  {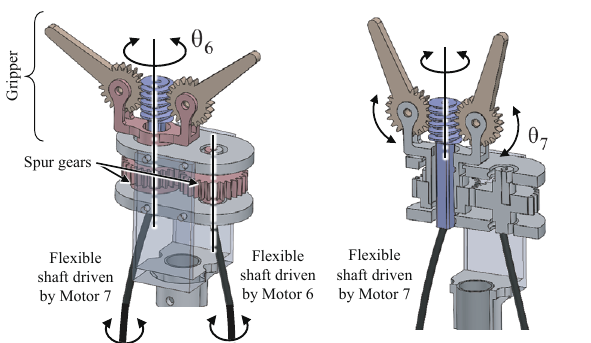}
  \caption{Design of coupled drive mechanism of gripper and J6.}
  \label{fig_gripper}
\end{figure}

\textbf{J6 and J7 (Gripper)}: 
Joints J6 and J7 are driven by a wrist-mounted coupled module as shown in Fig.~\ref{fig_gripper}, where 2-DoFs are generated through the relative motion of flexible shafts 6 and 7. For pure wrist yaw (J6), both shafts rotate at equal speed in opposite directions, driving the central worm gear and support frame to produce yaw rotation without gripper motion. For gripper actuation (J7), shaft 7 rotates while shaft 6 remains fixed, generating gripper actuation through a 10:1 worm reduction. In general, differential shaft rotation produces combined yaw and gripping motion.

\subsubsection{Manipulator Control Method}
In this study, no joint encoders are installed to minimize the moving mass of the arm section, and open-loop control is adopted. The joint angles are determined through inverse kinematics based on the manipulator link lengths. No compensation for the angular transmission error of the flexible shaft is implemented.

\section{Experiments}


To validate the proposed mechanism, we conducted experiments to answer the following research questions:

\begin{enumerate}
    \item Does the tensioner mechanism effectively suppress angular transmission error?
    \item Does the proposed method increase efficiency compared to conduit-covered transmission and Bowden cable?
    \item What performance characteristics does the manipulator exhibit during dynamic operation?
\end{enumerate}

\subsection{Experiment 1: Effects of Path Adjustment on angular transmission error and Efficiency}

\begin{figure}[t]
  \centering
  \includegraphics[width=0.9\linewidth]
  {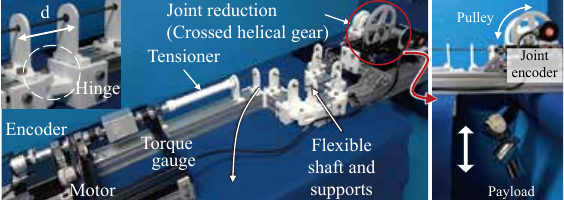}
  \caption{Setup for single-unit transmission experiment.}
  \label{fig_ex_single}
\end{figure}

\begin{figure}[t]
  \centering
  \includegraphics[width=1.0\linewidth]
  {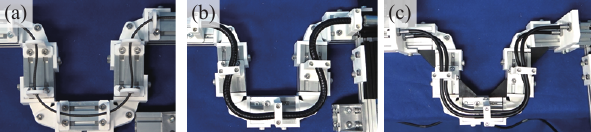}
  \caption{(a)~Flexible shaft with tensioner, (b)~Flexible shaft with conduit, (c)~Bowden cable.}
  \label{fig_configuration}
\end{figure}

To answer questions Q1: Does the tensioner suppress angular transmission error? and Q2: Does the proposed method increase the efficiency compared to other transmission methods? We prepared a single-unit experimental setup shown in Fig.~\ref{fig_ex_single} and measured the efficiency and angular transmission error of several remote actuation methods under bent configurations. This setup simulates the flexible shaft arrangement passing over multiple joints of an articulated manipulator, consisting of aluminium frame links connected by rotary hinges with the same length as the manipulator links, with flexible transmission modules mounted on top. A torque gauge (UTM III 10 Nm torque gauge, Unipulse Corp.) is installed between the motor and tensioner, and a pulley is attached to the backdrivable multi-start worm gearbox at the end part with output torque applied by winding up a wire suspending weights. The same aluminium frame links can also be equipped with (b)~conduit-covered flexible shafts and (c)~Bowden cables (Fig.~\ref{fig_configuration}). 
To maintain a consistent total bending angle across all transmission methods, configurations (b) and (c) are fixed with fewer supports than (a), as the stiffer outer tubes make it difficult to achieve sharp curvatures. Although (b) and (c) are designed to allow axial sliding of the Bowden cable and conduit-covered flexible shaft, no tensioning mechanism is incorporated; consequently, the effective bending angle of the tube is larger than that of the rigid links.
A comparison of the different transmission methods is listed in Table~\ref{tab:method_comparison}.

\begin{table}[t]
    \centering
    \small
    \caption{Comparison of mechanical parameters of each transmission method using experiment.}
    \label{tab:method_comparison}
    \begin{tabular}{lccc}
        \hline
        Method 
        & \makecell{Inner shaft \\ diameter} 
        & \makecell{Outer tube \\ diameter} 
        & \makecell{Joint reduction \\ ratio} \\
        \hline
        \makecell{W/ tensioner \\ (Proposed)} & 3~mm & None & 10:1 \\
        W/ conduit & 3~mm & 10~mm & 10:1 \\
        W/o tensioner & 3~mm & None & 10:1 \\
        Bowden cable & 3~mm & 6.5~mm & None (1:1) \\
        \hline
    \end{tabular}
\end{table}

Using this setup, we measured the efficiency and angular transmission error of several transmission methods. Efficiency is defined as follows: when the pulley outputs torque $\tau_p$ to lift the load while the motor applies input torque $\tau_m$ to the flexible shaft:

\begin{equation}
\text{Efficiency} = \frac{\tau_p}{\tau_m} = \frac{W R}{\tau_m i}
\end{equation}
where $W$, $R$, and $i$ are the weight of the payload, the outer radius of the pulley, and the reduction ratio of the joint, respectively. $\tau_m$ is measured by the torque gauge connected to the motor. Angular transmission error is defined as the error between the input and output angles, quantified by the rotation angle of the pulley when the load is applied at the joint with the input motor at rest. Two optical encoders employed at the motor and joint side measure these angles. 

An experiment was conducted to examine the effect of varying $d$ on performance characteristics using a flexible shaft of length 550~mm. The shaft was bent to $90^\circ$ at four locations along its length, as shown in Fig.~\ref{fig_configuration}~(a), to replicate the worst-case bending configuration. A constant external torque of 0.5~Nm with effective tensioner spring stiffness of 0.2~N/mm was used to evaluate slack length and efficiency, with the results of the experiment summarised in Table~\ref{tab:slack_efficiency}.

For $2d = 30$~mm, the system exhibits a lower efficiency of 51.3\% due to tighter bending and increased sliding friction between the shaft and bearings. When $2d = 70$~mm, the efficiency increases to 61.45\%; however, the angular transmission error rises to 9.4$^\circ$, and the larger slack length requires greater tensioner extension, resulting in a larger base size. In contrast, $2d = 50$~mm provides a balanced performance, with 60.93\% efficiency, moderate angular transmission error of 7.2$^\circ$, and a compact slack length of 55~mm. Therefore, considering efficiency, angular transmission error, and compactness, $2d = 50$~mm is selected as the optimal distance for the manipulator.

\begin{table}[t]
\centering
\caption{Slack, Efficiency, and Angular transmission error vs. Support Bearing Distance $d$.}
\label{tab:slack_efficiency}
\resizebox{\linewidth}{!}{%
\begin{tabular}{c c c c c c}
\hline
$2d$  & \multicolumn{3}{c}{Slack length [mm]} & Efficiency  & Angular transmission \\
\cline{2-4}
        [mm]  & $L_{\text{total}}$ & $L_{\max}$ & Measured & [\%] & error [deg] \\
\hline
30 & 25.75 & 35.15 & 35 & 51.30 & 6.0 \\
50 & 42.92 & 58.57 & 55 & 60.93 & 7.2 \\
70 & 60.08 & 82.01 & 77 & 61.45 & 9.4 \\
\hline
\end{tabular}%
}
\end{table}

\begin{figure}[t]
  \centering
  \includegraphics[width=1\linewidth]
  {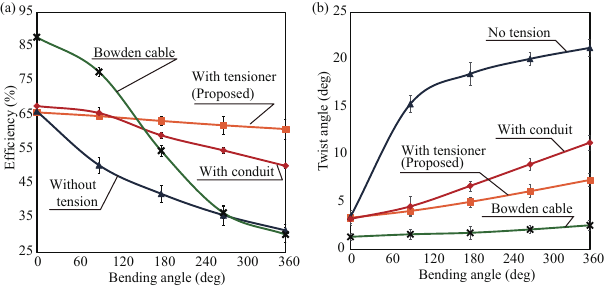}
  \caption{Result of bending experiment.}
  \label{fig_result_bending}
\end{figure}

In the experiments, we measure angular transmission error and efficiency for several transmission methods as the bending angle of each joint varies from 0$^{\circ}$ to 90$^{\circ}$ with applied torque of 0.5~Nm. The transmission methods tested include: flexible shaft with a tensioner (proposed), flexible shaft with an outer conduit, flexible shaft with a fixed sliding direction, and Bowden cable. The experimental results are shown in Fig.~\ref{fig_result_bending}~(a) and (b). The horizontal axis in both plots represents the total bending angle (four times the individual joint angle), while the vertical axes show (a) efficiency and (b) angular transmission error.
It should be noted that the total bending angle refers to the angle of the aluminium frame links, and the actual shaft or wire bending angle is larger than the frame angle for the Bowden cable, without tensioner, and with conduit configurations. This was unavoidable to maintain a consistent total frame bending angle across all methods.


First, the efficiency measurements indicate that all flexible shaft configurations exhibited approximately 66\% efficiency in the straight configuration. This baseline reduction is primarily attributed to losses within the joint reduction mechanisms rather than the transmission shafts themselves. As the total bending angle increased, efficiency decreased for all configurations. However, the proposed flexible shaft with tensioner demonstrated the smallest performance degradation, maintaining approximately 61\% efficiency even at a total bending angle of $360^{\circ}$. This represents the highest efficiency among all flexible shaft-based transmission methods tested.

The flexible shaft with outer conduit exhibited a similar trend to the proposed configuration, but its efficiency remained consistently about 10\% lower across the entire bending range. In contrast, the flexible shaft without a tensioner or conduit showed the lowest efficiency. Although torque transmission was still achievable, helical buckling of the shaft contributed to reduced efficiency and transmission instability. The Bowden cables exhibited the highest efficiency in straight configuration (87\%) and maintained superior performance up to a total bending angle of $90^{\circ}$. However, its efficiency decreased sharply to 33\% at $360^{\circ}$, indicating high frictional losses under larger curvatures. Hence, for applications requiring large bending angles, the proposed mechanism offers the most balanced performance in terms of efficiency retention and structural stability.

Regarding the angular transmission error results, all transmission methods exhibited an increase in angular transmission error as the bending angle increased. The Bowden cable demonstrated the smallest increase in angular transmission error, indicating the highest transmission precision among the evaluated methods. Among the flexible shaft transmission configurations, the proposed mechanism with a tensioning system exhibited the lowest angular transmission error, whereas the configuration without a tensioner showed the highest angular transmission error. Furthermore, it can be observed that increased angular transmission error leads to reduced transmission efficiency in flexible shaft-driven systems. Greater angular transmission error indicates larger torsional deformation under load, which results in energy loss, reduced positional accuracy, and decreased torque transmission efficiency. Therefore, minimizing angular transmission error is essential to ensure reliable and efficient power transmission in flexible shaft-driven multi-joint manipulators.

\subsection{Experiment 2: Evaluation of Manipulator Performance}

To evaluate the manipulator's performance, we conducted three experiments.

\begin{figure}[t]
  \centering
  \includegraphics[width=0.9\linewidth]
  {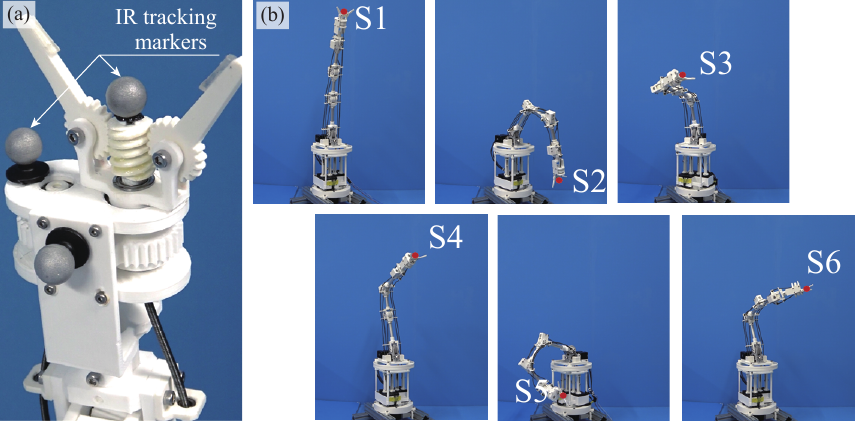}
  \caption{Transient motion of the manipulator to the predefined setpoints.}
  \label{fig_accuracy_test}
\end{figure}

\begin{table}[t]
    \centering
    \small
    \setlength{\tabcolsep}{4pt}
    \caption{Distance error of each checkpoint. (mean $\pm$ std) [mm]}
    \label{tab:distance_error}
    \begin{tabular}{lccccccc}
        \hline
         & S1 & S2 & S3 & S4 & S5 & S6 \\
        \hline
        \makecell[l]{Distance \\ Error [mm]}
        & \makecell{$3.28$ \\ $\pm 0.47$}
        & \makecell{$3.59$ \\ $\pm 1.65$}
        & \makecell{$3.06$ \\ $\pm 0.71$}
        & \makecell{$2.58$ \\ $\pm 1.14$}
        & \makecell{$1.65$ \\ $\pm 0.85$}
        & \makecell{$2.59$ \\ $\pm 1.51$}\\
        \hline
    \end{tabular}
\end{table}

\textbf{Repeatability}: We evaluated the trajectory tracking performance of the manipulator end-effector using an external motion capture system, as shown in Fig.~\ref{fig_accuracy_test}~(a). Reflective IR markers were attached to the manipulator tip, and its spatial position was recorded using a Motion capture (OptiTrack Prime x22 system, NaturalPoint Inc.) at a sampling frequency of 500~Hz. The manipulator was sequentially moved through predefined setpoints within its workspace as shown in  Fig.~\ref{fig_accuracy_test}~(b). 

The result is shown in Table~\ref{tab:distance_error}.  The coordinates at the first pass through each setpoint were used as the reference, and the mean and standard deviation of the distances between the reference and the coordinates at five subsequent passes through each setpoint were computed. The average error was approximately 1~mm to 4~mm for all setpoints with deviations less than 2~mm.  


\textbf{Moving Speed}: The dynamic performance of the manipulator was evaluated through a high-speed base rotation test with the manipulator maintained in a horizontal configuration. The base joint was commanded to rotate from $-180^\circ$ to $180^\circ$ within 1.3 seconds, corresponding to one full revolution at an average speed of 46.2~rpm. The position of the end-effector was recorded using the same setup shown in Fig.~\ref{fig_accuracy_test}~(a).

\begin{figure}[t]
  \centering
  \includegraphics[width=1\linewidth]
  {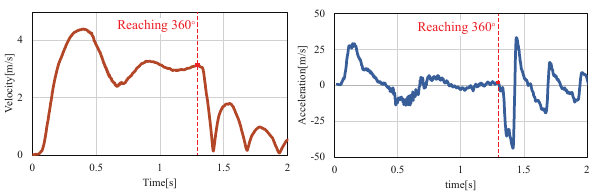}
  \caption{Result of moving speed experiment: (a)~Velocity, (b)~Acceleration of end-effector.}
  \label{fig_velocity_result}
\end{figure}

As shown in Fig.~\ref{fig_velocity_result}, we calculated the average velocity and average acceleration of the tip of the end-effector over 0.2-second intervals. The maximum velocity was 4.3~m/s, and the maximum acceleration was 39~m/s$^2$.


\begin{figure}[t]
  \centering
  \includegraphics[width=0.9\linewidth]
  {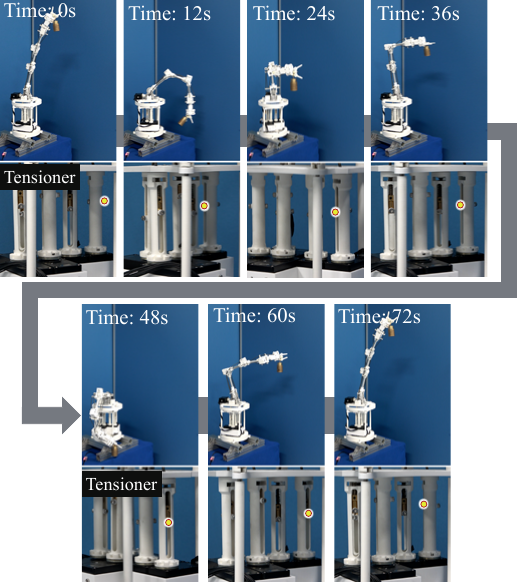}
  \caption{Motion of manipulator and tensioner carrying 500~g payload.}
  \label{fig_motion_with_payload}
\end{figure}

\textbf{Motion with 500~g Payload}: The manipulator was evaluated under load by suspending a 500~g payload at the end effector, as shown in Fig.~\ref{fig_motion_with_payload}. The figure illustrates the sequential motion of the manipulator while carrying the payload, with particular emphasis on the highlighted carriage of the tensioning mechanism. During operation, the manipulator moves smoothly while the carriage passively slides to accommodate length variations arising from simultaneous bending of multiple joints. This demonstrates that the proposed tensioning mechanism effectively maintains stable torque transmission during loaded manipulation.

\section{Discussion}

Unlike existing approaches that reinforce flexible shafts via conduits or cylindrical links, the proposed tensioner mechanism reduces angular transmission error while adding minimal mass to the arm section. The results from Experiment~1 demonstrate that the proposed system maintains over 60\% transmission efficiency even as bending angles increase, and achieves approximately 65\% lower angular transmission error compared to unsupported flexible shaft transmission, confirming its effectiveness as a remote actuation system. The lightweight (280~g arm section mass), slim, and compliant design minimizes injury risk during human contact, accelerating robot deployment in human environments.

To increase the current 500~g payload capacity, comprehensive optimization involving motor torque, shaft diameter, and reduction ratios is necessary. A critical challenge is that the restoring force of the bent flexible shaft becomes a load on the motors. Therefore, establishing a deformation model that accounts for material elasticity is essential for deriving design solutions that meet desired torque specifications.

The accuracy tests revealed positioning errors of approximately 3~mm, indicating sufficient capability for handling palm-sized objects. While the hardware demonstrates favorable characteristics, control methods that account for the manipulator's angular transmission error have not yet been established. Two potential approaches include closed-loop control using joint encoders or learning-based control methods~\cite{habich2025generalizable} that incorporate the physical characteristics of the flexible shaft transmission.

\section{Conclusion}


This paper focuses on flexible-shaft remote transmission to reduce arm-section mass, friction losses, and angular transmission errors through passive sequential optimization of the path geometry using a tensioner mechanism. The proposed manipulator features a lightweight design, with an arm section mass of 280~g, corresponding to 60\% of the 500~g payload capacity. Experimental results demonstrate that transmission efficiency is maintained over 60\% even at a cumulative bending angle of 360$^{\circ}$. The angular transmission error of 7.3$^{\circ}$ is smaller than both that of a flexible shaft without tensioner, and that of a conduit-covered flexible shaft, and the manipulator accuracy test results showed approximately 3~mm accuracy. 
This approach enables lightweight and safe operation of multi-joint, multi-DoF manipulators with payload capacities in the order of several hundred grams, where positioning errors of several millimeters are acceptable. The slim configuration reduces the risk of damaging people or objects during unexpected collisions in cluttered environments with high collision probability, enabling operation in settings where conventional rigid robots would pose significant safety hazards.

Future work will incorporate encoder feedback or model-based compensation to improve joint angle control accuracy under loads~\cite{rodriguez2019flexible}. 
Also, we will evaluate safety in tasks involving collisions and contact-based manipulation, leveraging the inherent angular compliance of the proposed design.




\bibliographystyle{IEEEtran}
\bibliography{reference}

\end{document}